\documentclass[9pt,conference]{IEEEtran}
\IEEEoverridecommandlockouts
\usepackage{cite}
\usepackage{amsmath,amssymb,amsfonts,amsthm}
\usepackage{algorithm}
\usepackage{algorithmicx}
\usepackage{algpseudocode}

\usepackage{graphicx}
\usepackage{textcomp}
\usepackage{xcolor}
\usepackage{hyperref}
\hypersetup{hypertex=true}
\def\BibTeX{{\rm B\kern-.05em{\sc i\kern-.025em b}\kern-.08em
    T\kern-.1667em\lower.7ex\hbox{E}\kern-.125emX}}
\usepackage{makecell}
\usepackage{./tabularray}

\theoremstyle{definition}

\begin{document}

\newcommand{\ml}[1]{{\color{red}\bf [Meng: #1]}}
\newcommand{\red}[1]{{\color{red}\bf (#1)}}
\newcommand{\ling}[1]{{\color{violet}\bf [Ling: #1]}}

\title{\textbf{HybriMoE: Hybrid CPU-GPU Scheduling and Cache Management for Efficient MoE Inference}}

\author{\IEEEauthorblockN{Shuzhang Zhong$^{1,2}$, Yanfan Sun$^{5}$, Ling Liang$^{2}$, Runsheng Wang$^{2,3,4}$, Ru Huang$^{2,3,4}$, Meng Li$^{1,2,4*}$}
\IEEEauthorblockA{$^1$Institute for Artificial Intelligence, Peking University, Beijing, China}
\IEEEauthorblockA{$^2$School of Integrated Circuits, Peking University, Beijing, China}
\IEEEauthorblockA{$^3$Institute of Electronic Design Automation, Peking University, Wuxi, China}
\IEEEauthorblockA{$^4$Beijing Advanced Innovation Center for Integrated Circuits, Beijing, China}
\IEEEauthorblockA{$^5$School of Computer Science and Engineering, Beihang University, Beijing, China}

\thanks{
This work was supported in part by NSFC under Grant 62495102 and Grant 92464104, in part by the National Key Research and Development Program under Grant 2024YFB4505004, in part by Beijing Municipal Science and Technology Program under Grant Z241100004224015, and in part by 111 Project under Grant B18001.

$^*$Corresponding author: meng.li@pku.edu.cn}
}

\maketitle
\begin{abstract}

The Mixture of Experts (MoE) architecture has demonstrated significant advantages as it enables to increase the model capacity without a proportional increase in computation. However, the large MoE model size still introduces substantial memory demands, which usually requires expert offloading on resource-constrained platforms and incurs significant overhead. Hybrid CPU-GPU inference has been proposed to leverage CPU computation to reduce expert loading overhead but faces major challenges: on one hand, the expert activation patterns of MoE models are highly unstable, rendering the fixed mapping strategies in existing works inefficient; on the other hand, the hybrid CPU-GPU schedule for MoE is inherently complex due to the diverse expert sizes, structures, uneven workload distribution, etc. To address these challenges, in this paper, we propose HybriMoE, a hybrid CPU-GPU inference framework that improves resource utilization through a novel CPU-GPU scheduling and cache management system. HybriMoE introduces (i) a dynamic intra-layer scheduling strategy to balance workloads across CPU and GPU, (ii) an impact-driven inter-layer prefetching algorithm, and (iii) a score-based caching algorithm to mitigate expert activation instability. We implement HybriMoE on top of the kTransformers framework and evaluate it on three widely used MoE-based LLMs. Experimental results demonstrate that HybriMoE achieves an average speedup of 1.33$\times$ in the prefill stage and 1.70$\times$ in the decode stage compared to state-of-the-art hybrid MoE inference framework. Our code is available at: \href{https://github.com/PKU-SEC-Lab/HybriMoE}{https://github.com/PKU-SEC-Lab/HybriMoE}.


\end{abstract}

\section{Introduction}
\label{sec:intro}

Mixture of Experts (MoE) has emerged as a promising solution to enhance computational efficiency of Large Language Models (LLMs) without compromising model performance\cite{shazeer2017outrageously, masoudnia2014mixture}. By employing dynamic routing functions that allocate input tokens to a subset of experts, MoE enables the scaling of LLM parameters and capabilities without a proportional increase in computational demands.

Despite its advantages, MoE introduces significant memory requirements, which pose a particular challenge for deployment on edge devices with limited memory resources. To mitigate this issue, expert offloading techniques store expert weights in secondary storage, such as CPU memory or SSDs, and load them into GPU memory through PCIE on demand\cite{eliseev2023fast}. In such offloading scenarios, the primary bottleneck becomes the overhead associated with on-demand loading, driven by the large communication scale and limited bandwidth. To mitigate this problem, several studies have explored quantization, prefetching or caching techniques to reduce latency\cite{hwang2023pre,zhong2024adapmoe,song2024promoe,tang2024hobbit}.

\begin{figure}
    \centering
    \includegraphics[width=\linewidth]{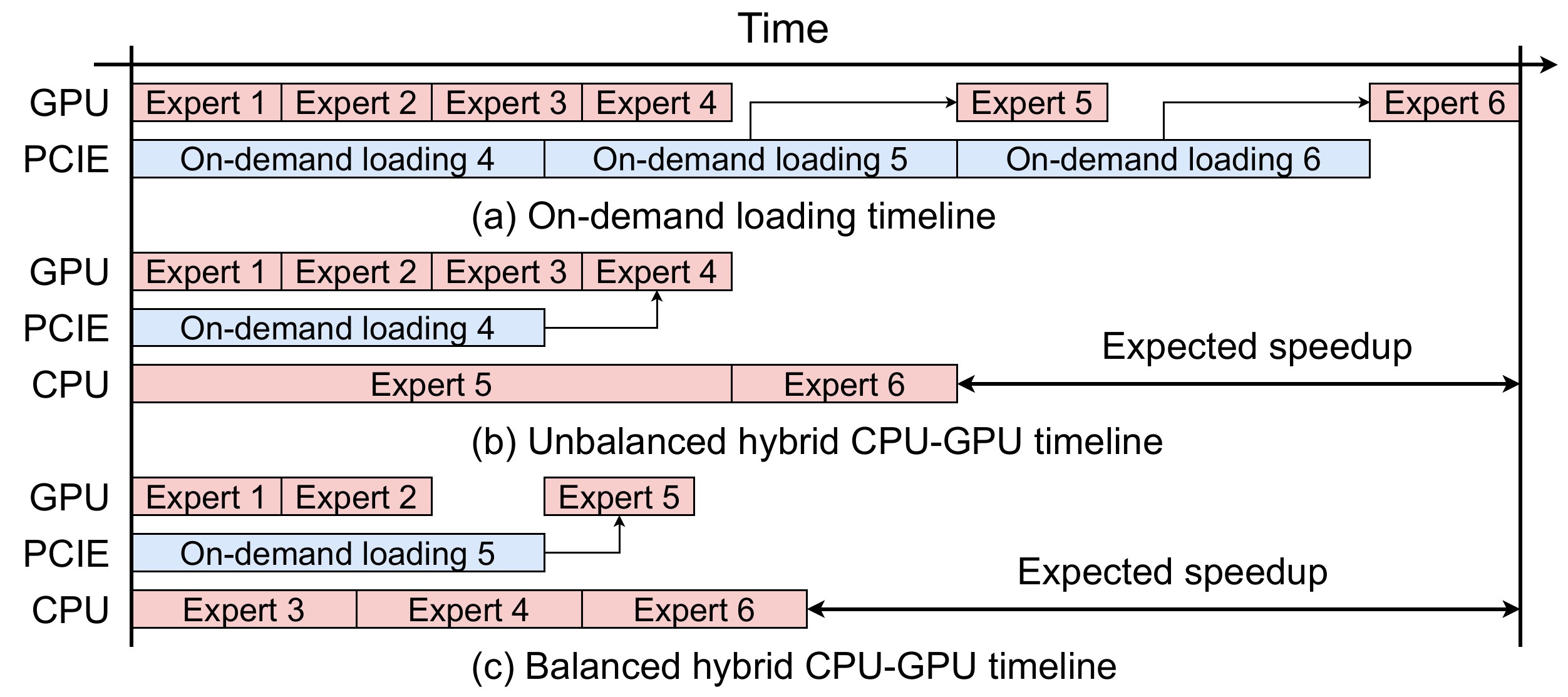}
    \caption{Execution timeline of three scenarios. Expert computation time on the GPU remains constant, while CPU execution time increases linearly with workload. The balanced scheduling in (c) achieves improved utilization and reduces overall execution time.}
    \label{fig:background}
\end{figure}

Previous works in other offloading scenarios have further explored leveraging CPU computation to reduce the frequency of memory transfers\cite{you2021ship,park2024improving}. Techniques such as PowerInfer\cite{song2023powerinfer} and Caraserve\cite{li2024caraserve} have achieved notable success by exploiting activation patterns or optimizing adapter usage during inference. Similarly, MoE-specific offloading approaches, including Fiddler\cite{kamahori2024fiddler} and kTransformers\cite{ktransformers}, utilize the CPU to execute expert layers during cache misses. As illustrated in Figure~\ref{fig:background}, when a cache miss occurs, the CPU processes the corresponding expert computation instead of transferring the layer to the GPU, reducing data transfer overhead.

While CPU computation is effective for traditional inference tasks, MoE models present unique challenges that complicate their application. Expert activations in MoE models are typically less skewed and exhibit significant variability across iterations, making it difficult to predict which experts will be activated\cite{song2024promoe}. This dynamic behavior complicates the balancing of workloads between CPU and GPU, as static task allocation strategies fail to adapt to real-time changes in workload distribution. However, existing solutions rely on \textbf{fixed mapping strategies} based on historical activation frequencies, neglecting the dynamic and unpredictable nature of MoE inference. These limitations result in suboptimal resource utilization and increased inference latency as illustrated in figure~\ref{fig:background}(b) and (c).

In light of these challenges and opportunities, we propose \textbf{HybriMoE}, a hybrid CPU-GPU scheduling and cache management system to improve the efficiency of MoE inference. 
Reducing latency in hybrid systems requires maximizing hardware resource utilization, which depends on effective task-to-hardware mapping. However, the dynamic nature of MoE models poses significant challenges to designing optimal mapping strategies.
To address this, HybriMoE introduces a comprehensive optimization framework to improve mapping efficiency through three key directions: (i) intra-layer hybrid scheduling, (ii) inter-layer prefetching, and (iii) inter-iteration cache management.
The key contributions of HybriMoE are as follows:

\begin{itemize}
    \item \textbf{Hybrid MoE CPU-GPU Scheduling.} An efficient hybrid scheduling algorithm for MoE inference that dynamically balances workloads across GPUs and CPUs, optimizing resource utilization and minimizing latency through prioritized task execution and data transfer management.
    \item \textbf{Impact-driven prefetching.} A prefetching mechanism that simulates the potential impact of preloading experts from subsequent layers and prioritizes those with the higher expected gains.
    \item \textbf{MoE-specialized Cache Management.} An expert score-based caching strategy that prioritizes high-demand experts across layers to minimize cache misses.
    \item \textbf{System Implementation.} We implement HybriMoE on top of ktransformers framework. We evaluate HybriMoE on three popular MoE-based LLMs and various platforms. Compared to existing hybrid scheduling methods, HybriMoE achieves 1.33$\times$ and 1.70x speedup on prefill and decode stages respectively. 
\end{itemize}


\section{Background}

\subsection{Mixture-of-Experts}
Mixture-of-Experts (MoE) models offer an efficient solution for handling the computational demands of LLMs by activating only a subset of experts\cite{masoudnia2014mixture,fedus2022switch,costa2022no,yang2024qwen2}. Unlike traditional dense networks, MoE models use a gating function $G$ to select which experts process a given input token, reducing the number of active parameters and improving computational efficiency. Given an input $x$ and N experts $E_{0},...E_{N-1}$ the output $y$ of the MoE layer can be expressed as:
\begin{align}
    y =\sum_{i=0}^{N-1} \text{Softmax}(\text{TopK}(x\cdot W_g))_i   E_i(x)
\end{align}

\begin{figure}
    \centering
    \includegraphics[width=0.95\linewidth]{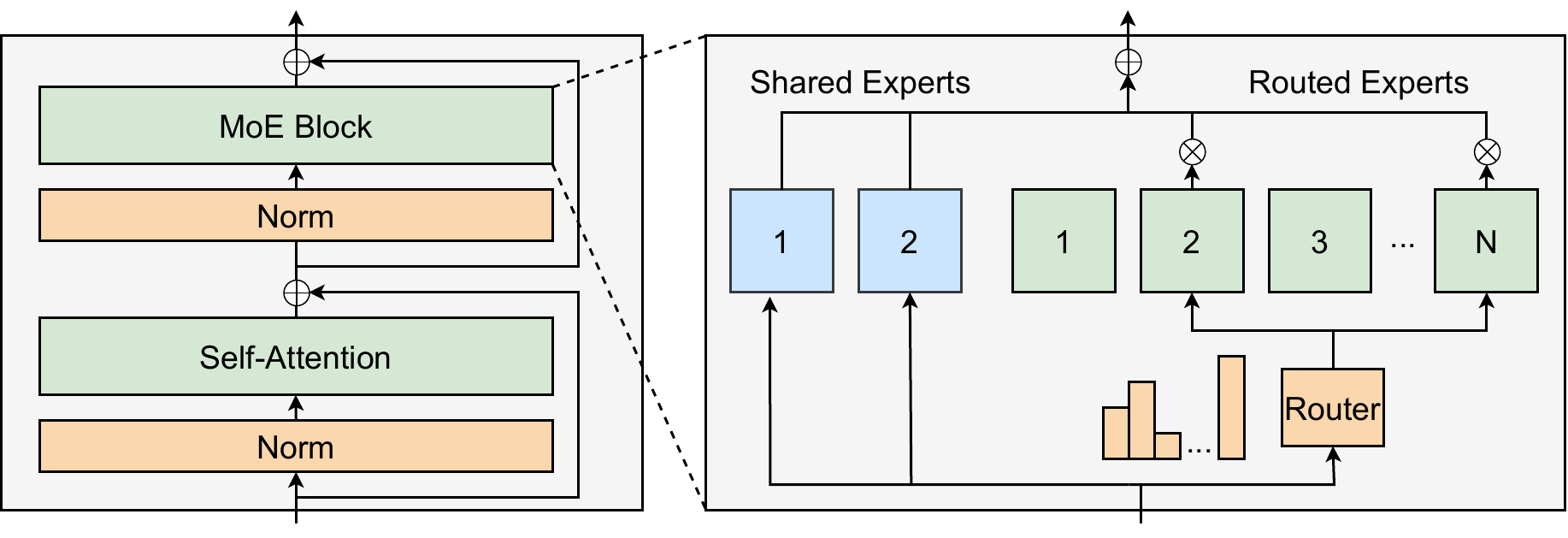}
    \caption{An example of MoE architecture with shared and routed experts. }
    \label{fig:moe}
\end{figure}

The total number of experts N and the number of activated experts K vary among different MoE implementations. For instance, the Mixtral model employs 8 experts, with only 2 being active at a time\cite{jiang2024mixtral}.
In contrast, DeepSeek utilizes 64 experts, activating 6 at once. This larger, finer-grained expert pool allows for greater specialization and more efficient knowledge acquisition\cite{liu2024deepseekv2,dai2024deepseekmoe}.
Additionally, as shown in figure \ref{fig:moe}, DeepSeek employs a shared expert strategy, where a subset of experts—known as shared experts—are activated for all tokens. The original experts are defined as routed experts. This reduces redundancy among experts, ensuring efficient processing by minimizing unnecessary computational overlap, thus enhancing overall model efficiency.
\subsection{Efficient MoE Offloading}
Parameter-offloading techniques have been proposed to address the significant memory requirements of large language models (LLMs) \cite{aminabadi2022deepspeed,sheng2023flexgen}. However, these techniques are primarily designed for dense models and involve loading or prefetching all parameters, leading to unnecessary communication overhead. To accommodate the sparse activation patterns in Mixture-of-Experts (MoE) models, several specialized techniques have been introduced, including advanced gating, prefetching, and quantization strategies \cite{zhong2024adapmoe,li2023adaptive,song2024promoe,wei2024aptmoe,hwang2023pre,tang2024hobbit,eliseev2023fast,li2024examining,xue2024moe,he2024expertflowoptimizedexpertactivation}. These methods aim to minimize the on-demand loading overhead, reducing unnecessary memory transfers and improving overall performance.

\subsection{Hybrid CPU-GPU Scheduling}

Previous offloading techniques have primarily focused on reducing memory transfer overhead by offloading certain computations to the CPU\cite{park2024improving}. For instance, PowerInfer \cite{song2023powerinfer} reduces GPU memory demand by executing less frequently activated neurons on the CPU, taking advantage of skewed activation patterns. Caraserve \cite{li2024caraserve} addresses cold-start delays in LoRA serving by utilizing CPU assistance and employing rank-aware scheduling to reduce latency. These methods are effective in scenarios where activations are skewed or tasks have long periods of parameter reuse.

In the context of MoE models, techniques like Fiddler\cite{kamahori2024fiddler} and kTransformers\cite{ktransformers} extend this concept by offloading expert layer computation to the CPU during cache misses. Specifically, when an expert is not in the GPU cache, the CPU executes the corresponding expert layer instead of loading it from memory. These approaches aim to optimize memory usage by exploiting CPU-GPU parallelism and mitigating the overhead of loading large models onto the GPU.

In table~\ref{tab:compare}, we compare HybriMoE with prior-art works qualitatively. As can be observed, HybriMoE features CPU-GPU hybrid scheduling to improve the efficiency of both prefill and decode stages.
\begin{table}
    \centering
\caption{Comparison with existing offloading works.}
\label{tab:compare}
    \begin{tabular}{c|c|c|c|c} \hline \hline
         &  \makecell{Offload\\Granularity}&  \makecell{CPU\\Computation}&  \makecell{Dynamic\\Mapping}& \makecell{Cache\\Optimization}\\ \hline
 Powerinfer& Neuron& Decode& $\checkmark$&LFU\\ 
      
 llama.cpp& Layer& Prefill+Decode& $\times$&LFU\\   AdapMoE&  Expert&  $\times$&  $\checkmark$& LRU\\  
         KTrans&  Expert&  Decode&  $\times$& LFU\\ \hline 
         Ours&  Expert&  Prefill+Decode&  $\checkmark$& Score-Aware\\\hline \hline
    \end{tabular}
\end{table}

\section{Motivation}
\label{sec:motivation}

The primary bottleneck in existing hybrid CPU-GPU scheduling for MoE inference is the \textbf{suboptimal resource utilization}.
To address this issue, we begin by analyzing the main challenges of finding an efficient \textbf{mapping strategy}.

\begin{figure}
    \centering
    \includegraphics[width=\linewidth]{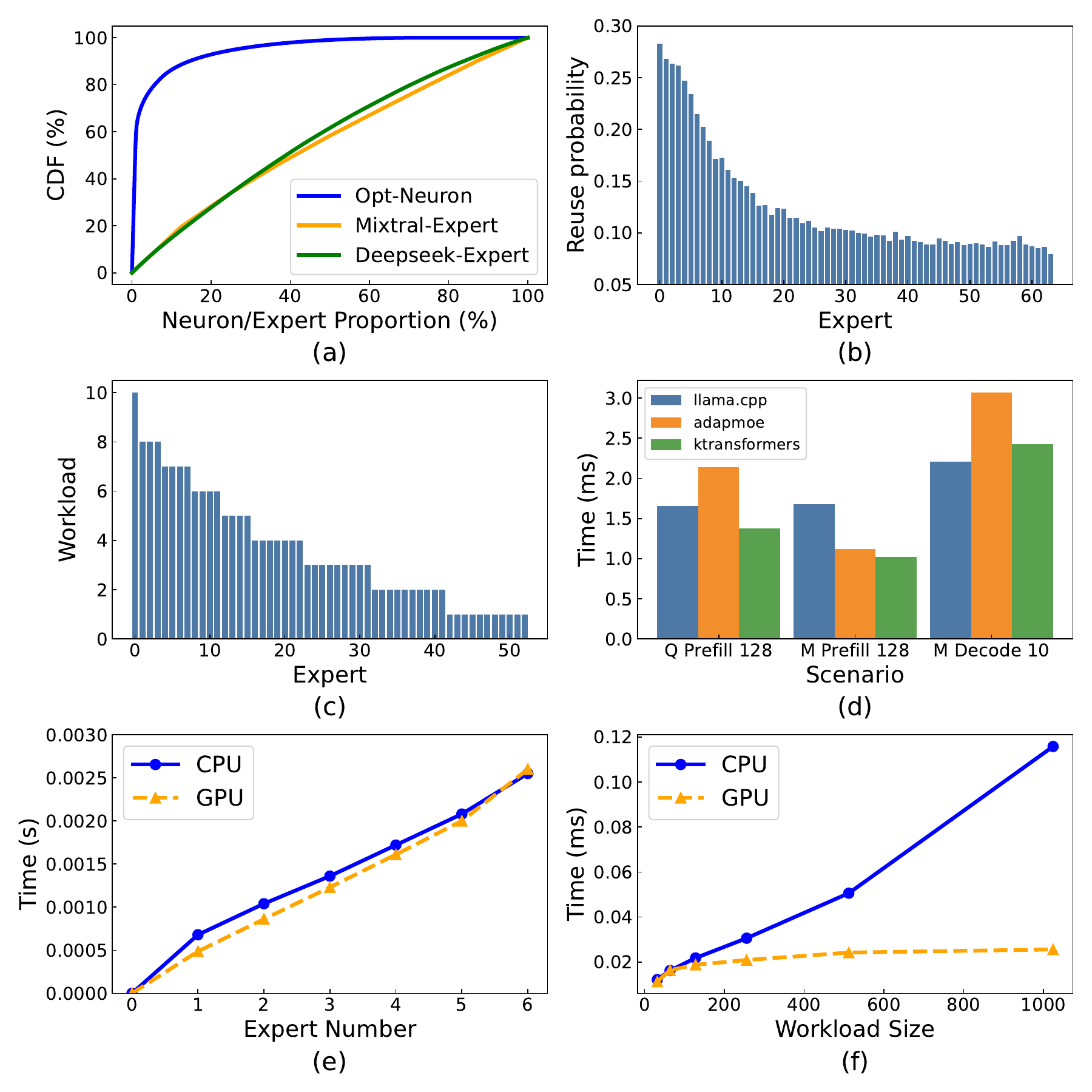}
    \caption{
    (a) Cumulative activation frequency(CDF) for neurons and experts,
    (b) Reuse probability of experts by score, suggesting cache optimization opportunities, 
    (c) Expert workload distribution of DeepSeek in a prefill forward,
    (d) Latency of prefill 128 tokens for Qwen2(Q), Mixtral(M) and decode 10 tokens for Mixtral with three existing methods,
    (e) CPU vs. GPU time for varying numbers of experts at fixed load, with CPU benefiting from overlapping computations. (f) CPU and GPU time across workload sizes.}
    \label{fig:motivation}
\end{figure}

\textbf{Challenge 1: High Instability of MoE Activation Patterns.} In existing hybrid CPU-GPU scheduling research, both sparse models with highly skewed activations, like PowerInfer, and dense models (or LoRA inference) exhibit relatively stable activation patterns. In these models, activation is either concentrated on a few `hot' neurons or remains consistent over time, making scheduling and workload balancing easier.
In contrast, MoE models have unpredictable activation patterns, with experts being activated in a dynamic and frequently changing manner. As shown in figure~\ref{fig:motivation}(a), compared with neuron-level sparsity, the activation frequency of MoE is more evenly distributed, making it challenging to predict the future expert usage. This lack of stability makes it difficult to determine an optimal CPU-GPU scheduling strategy in advance, leading to suboptimal resource utilization and inefficiency.

\textbf{Opportunity 1: MoE-specific Cache and Prefetch Optimization.} 
Despite the instability of MoE activations, certain temporal and structural patterns present opportunities for optimization. The temporal correlation of expert activation provides a basis for cache optimization: experts with higher activation scores are more likely to be reused in the next iteration as shown in Figure~\ref{fig:motivation}(b), suggesting that retaining high-score experts in cache can reduce access latency. Additionally, MoE models often exhibit high activation similarity between adjacent layers, which can be leveraged for prefetching. These MoE-specific optimizations provide a promising approach to reducing the challenges posed by the dynamic nature of expert activation.

\textbf{Challenge 2: Complexity of MoE Structure and Dynamic Scheduling.}
Minimizing latency in MoE inference requires maximizing hardware utilization, but existing fixed-mapping methods often lead to load imbalances and underutilized resources. The scheduling complexity is further increased by the diverse structures of MoE models, with variations in shared expert usage, expert size and number, and runtime cache behavior. Additionally, uneven load distribution and variable execution order in the prefill stage make efficient scheduling even more challenging as shown in figure~\ref{fig:motivation}(c). Given the need for layer-by-layer adjustments, a static optimal solution is impractical, making real-time scheduling a significant challenge. As illustrated in figure~\ref{fig:motivation}(d), the performance of three existing strategies vary in different stages and models. 

\textbf{Opportunity 2: MoE-specific scheduling rules.}
Despite the NP-Hard nature of the scheduling problem, in the specific context of MoE inference on CPU-GPU systems, several key observations can guide the design of efficient scheduling rules. First, expert transfer times remain relatively constant, simplifying decision-making. Additionally, GPU computation time scales linearly with the number of activated experts, while CPU computation benefits from overlapping memory access and computation due to its larger cache. As shown in Figure~\ref{fig:motivation}(e), the first expert computation on the CPU is slower, but subsequent tasks are processed faster with better cache utilization. Similarly, Figure~\ref{fig:motivation}(f) shows that GPU time remains stable with increasing workload, whereas CPU time grows linearly with workload. Leveraging these patterns, predefined scheduling rules can help achieve efficient workload balancing for MoE models.

\section{HybriMoE Design}
\subsection{Overview}
\begin{figure}
    \centering
    \includegraphics[width=0.9\linewidth]{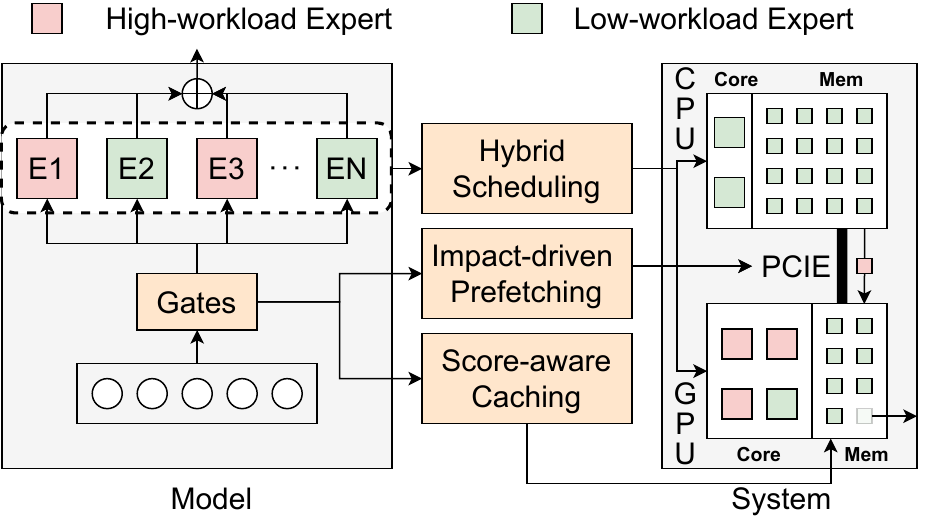}
    \caption{Overview of HybriMoE.}
    \label{fig:overview}
\end{figure}
This paper introduces HybriMoE, a CPU-GPU hybrid scheduling system tailored for MoE inference on memory-limited devices. HybriMoE addresses the challenges of unbalanced hardware utilization caused by the dynamic activation patterns and structural complexity of MoE models. The system incorporates three key techniques: (i) an efficient hybrid scheduling algorithm that dynamically distributes workloads between GPUs and CPUs, (ii) a score-based expert caching strategy that prioritizes high-demand experts to minimize cache misses, and (iii) an impact-driven prefetching mechanism that predicts and preloads high-demand experts, further enhancing resource utilization and reducing latency.

Figure~\ref{fig:overview} illustrates the overview of HybriMoE. The system begins with a warmup phase to collect essential performance metrics, such as CPU and GPU processing speeds and data transfer latency. During inference, HybriMoE leverages this information to implement hybrid CPU-GPU scheduling, score-aware caching, and impact-driven prefetching, ensuring efficient task execution and optimized resource usage throughout the inference process.

\subsection{Hybrid Scheduling Strategy}
\label{schedule}
The scheduling problem in MoE inference is inherently complex due to the dynamic nature of expert activation and the need to balance workloads across heterogeneous resources. To address these challenges, HybriMoE proposes a hybrid scheduling strategy that simplifies the task-to-hardware mapping by introducing three key priority rules:

\begin{itemize}
    \item \textbf{GPU Priority}: The GPU prioritizes the computation of cached experts, executing higher-load experts first.
    \item \textbf{CPU Priority}: The CPU prioritizes the computation of uncached experts, focusing on lower-load tasks for efficient execution. Additionally, the CPU can process cached experts when the CPU is idle, following a low-to-high load order.
    \item \textbf{Transfer Priority}: The CPU-GPU transfer mechanism prioritizes the movement of high-load uncached experts from CPU to GPU to minimize computation delays.
\end{itemize}

These rules constrain the ordering of experts on devices, simplifying the scheduling problem into an allocation problem:

\begin{align}
    \mathop{\arg\min}\limits_{cpu\_expert, gpu\_expert} \max(CPU_{TIME}(cpu\_expert), \nonumber
    \\ GPU_{TIME}(gpu\_expert))
\end{align}
This formulation does not account for the finish time of data transfers, as expert loading must be completed before GPU computation begins.

Based on these priority rules, HybriMoE divides all activated experts into a GPU queue and a CPU queue. The GPU queue contains cached experts on the GPU, sorted by load in descending order. The CPU queue contains uncached experts on the CPU, sorted by load in ascending order.

Before the actual execution, HybriMoE performs a simulation phase to evaluate scheduling strategies and identify an efficient task allocation plan tailored to the specific workload. This simulation approximates the execution process by iteratively filling the CPU computation, GPU computation, and data transferring timelines, enabling the system to determine a scheduling configuration that minimizes overall latency while balancing resource utilization across heterogeneous hardware.

 During each step of the simulation, the system selects the timeline with the earliest completion time and executes the corresponding operation—either a computation task on the CPU or GPU, or a data transfer via PCIE. Task selection adheres to the scheduling priorities: the GPU prioritizes high-load cached experts, the CPU focuses on low-load uncached experts and, when its queue is empty, processes low-load cached experts from the GPU queue, while PCIE prioritizes high-load uncached experts for faster availability on the GPU.

If an expert is transferred from the CPU to the GPU, it is inserted into the GPU queue in descending order of load, ensuring high-load tasks are prioritized for GPU computation. This iterative simulation continues until all experts are computed, effectively modeling the execution process and testing different scheduling strategies.

\begin{figure}
    \centering
    \includegraphics[width=\linewidth]{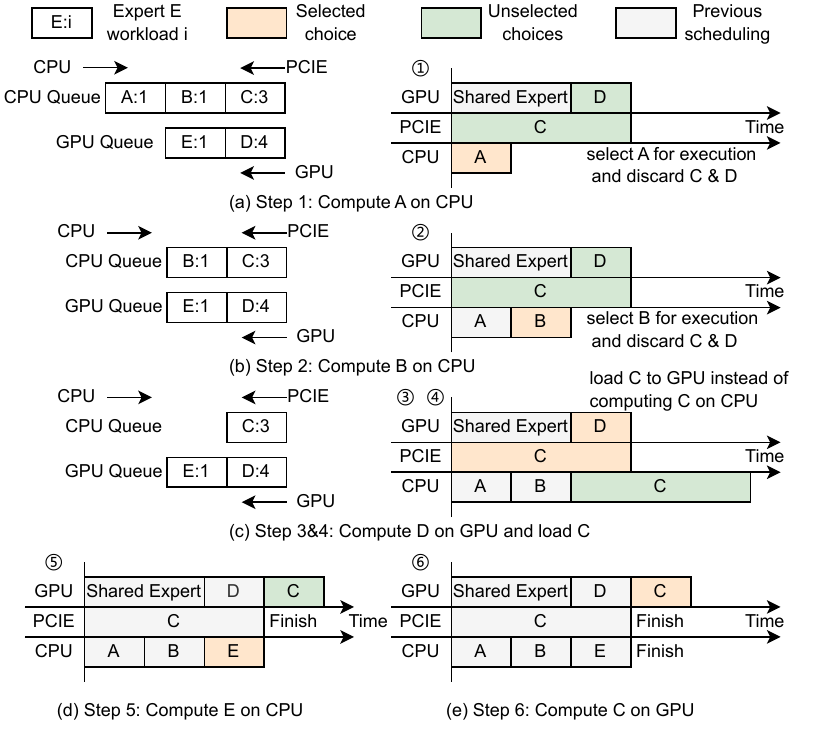}
    \caption{An example of hybrid scheduling. The CPU computes the cached expert E while the GPU computes the uncached expert C to achieve better hardware utilization.}
    \label{fig:schedule_example}
\end{figure}

The scheduling process is illustrated through an example in figure~\ref{fig:schedule_example}.
In this scenario, the GPU computation time is assumed to be constant, the CPU computation time is proportional to the expert's load, and the transmission time is fixed at 3 units. The scheduling algorithm in HybriMoE identifies an optimal strategy where the CPU computes the cached expert E while the GPU processes the uncached expert C, effectively improving hardware utilization by balancing workloads across resources.

\subsection{Impact-driven prefetching}
Due to the residual connections in LLMs, hidden states across consecutive layers exhibit a high degree of similarity, making expert prefetching an effective method to optimize resource utilization. While several existing works have adopted prefetching mechanisms, none of them discuss the critical trade-offs involved when multiple subsequent layers' experts can be prefetched. Specifically, these works do not explore how to strategically decide which layer’s experts should be prioritized for prefetching to maximize resource efficiency.

\begin{figure}
    \centering
    \includegraphics[width=\linewidth]{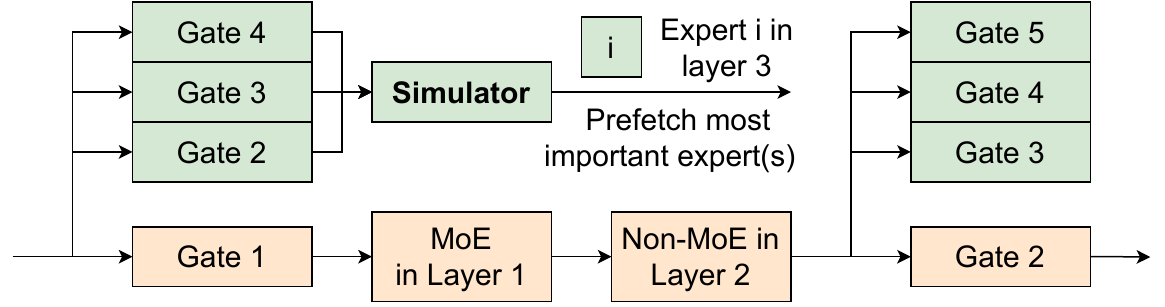}
    \caption{Impact-driven prefetch workflow.}
    \label{fig:prefetch}
\end{figure}

Inspired by the scheduling algorithm in \ref{schedule}, we propose impact-driven prefetching. Before executing a prefetch, the system performs a simulation to evaluate the potential gains of prefetching specific experts. This simulation estimates the impact of preloading a given expert on overall scheduling efficiency, allowing the algorithm to prioritize experts that yield the highest resource utilization improvements. The greedy nature of this simulation ensures minimal computational overhead, making it practical for real-time inference scenarios.

Specifically, HybriMoE predicts expert activations for the next three layers by reusing the gating information from those layers as illustrated in figure~\ref{fig:prefetch}. This prediction guides the prefetching mechanism, enabling the system to efficiently preload experts likely to be activated in subsequent computations. 

\subsection{Score-aware Caching}

Traditionally, the Least Frequently Used (LFU) and Least Recently Used (LRU) algorithms have been employed for MoE cache management. However, these strategies fail to account for the specific activation patterns observed in MoE models, where expert scores provide valuable predictive signals for future activations. As discussed in \ref{sec:motivation}, not only are currently activated experts likely to be reused in the future, but experts with high scores that were not activated also exhibit a higher probability of being selected in subsequent iterations.

To leverage this insight, we propose Score-Aware Caching, a novel cache replacement strategy tailored for MoE models. Specifically, we introduce the \textbf{Minus Recent Score (MRS)} replacement policy, which prioritizes retaining experts based on their routing scores.

Define s as the routing scores of all experts in the current iteration, S as the estimated priority score, $\alpha$ as the averaging coefficient, the update of the estimated priority can be expressed as :
\begin{align}
    S = \alpha\times TopP(s) + (1-\alpha) \times S
\end{align}
Here, only the top p expert scores will be accumulated. This is derived from the observation in figure~\ref{fig:motivation}(b) that the reuse probability of experts with lower scores does not exhibit significant differences. Typically, we set p to twice the number of activated experts.

\section{System Implementation}

We implement the HybriMoE system on top of the kTransformers framework and llama.cpp kernels. KTransformers provides a flexible infrastructure for kernel injection, enabling seamless support for hybrid CPU-GPU execution. To optimize the system workflow, we incorporate parallel execution across CPU, GPU, and PCIe transfers, utilizing fine-grained CUDA stream scheduling for efficient resource management. Additionally, we modify the C++ kernels to handle expert computation task allocation directly, minimizing redundant Python overhead and improving execution efficiency.
For quantization, we leverage Marlin quantization, a state-of-the-art 4-bit quantization kernel from llama.cpp, to significantly enhance computational efficiency and reduce memory usage.

\begin{figure*}[!tb]
    \centering
    \includegraphics[width=\linewidth]{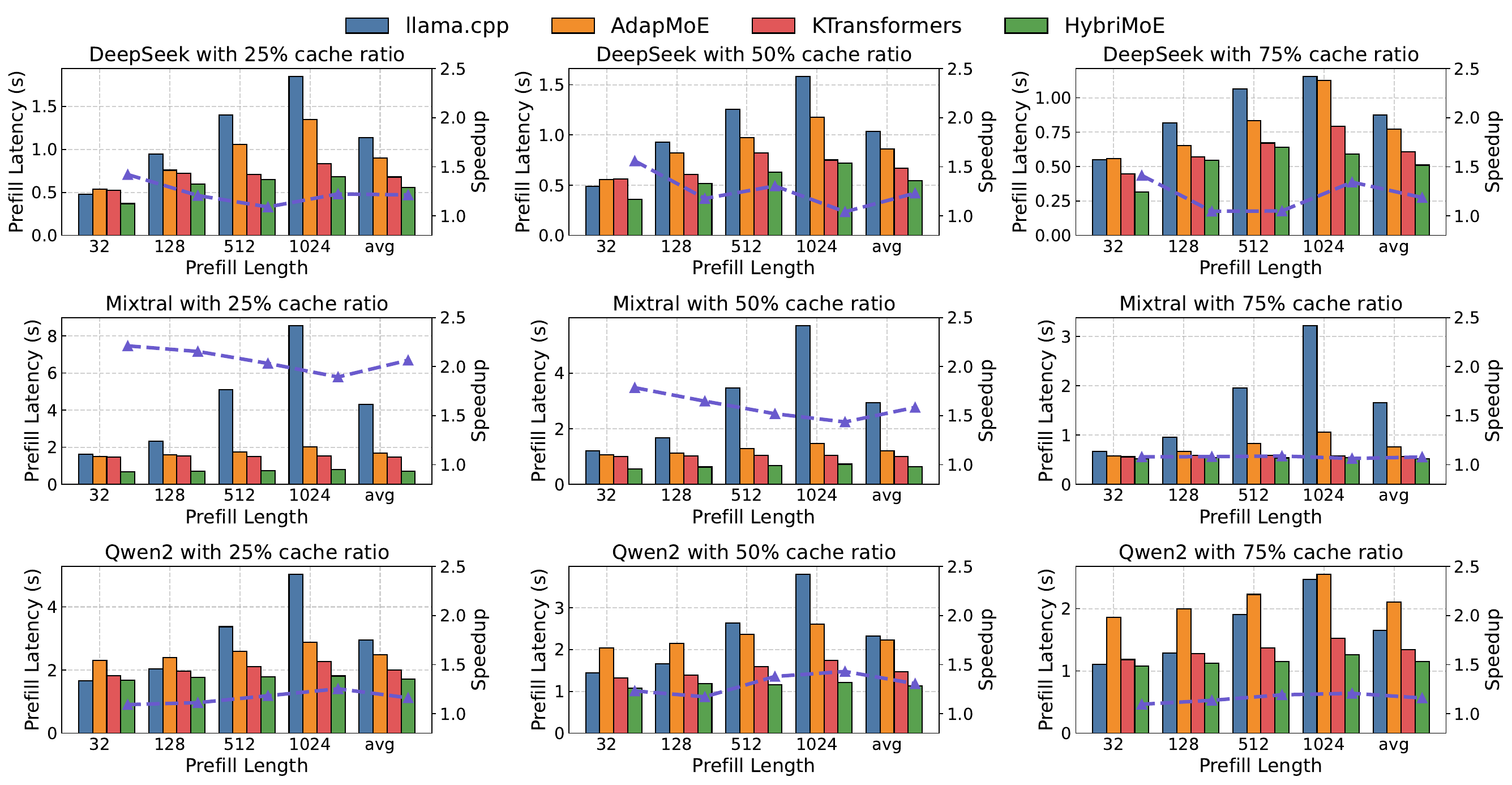}
    \caption{Prefill stage performance comparison across different input lengths and cache ratios, highlighting relative speedups against kTransformers.}
    \label{fig:prefill}
\end{figure*}
\begin{figure*}
    \centering
    \includegraphics[width=\linewidth]{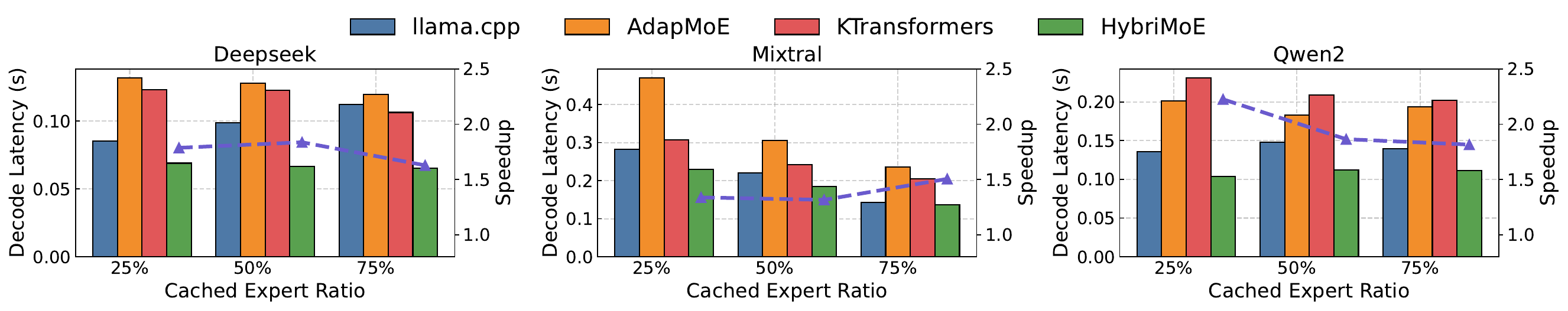}
    \caption{Decode stage performance comparison across different cache ratios.}
    \label{fig:decode}
\end{figure*}
\section{Experimental Results}

\subsection{Experimental Setup}
\subsubsection{Platforms}
We evaluate HybriMoE on the NVIDIA RTX A6000. For the CPU, we utilize an Intel Xeon Gold 5220R processor, restricting usage to 10 cores to simulate real-world edge deployment scenarios with limited resources.
To assess the system's performance and scalability under varying hardware configurations, we adjust the upper bound of the GPU expert cache ratio.

\subsubsection{Models}
We evaluate our system using three widely adopted MoE models with distinct characteristics: Mixtral-8x7B-Instruct\cite{jiang2024mixtral} (Mixtral), DeepSeek-V2-Lite-Chat\cite{liu2024deepseekv2} (DeepSeek), and Qwen2-57B-A14B-Instruct\cite{yang2024qwen2} (Qwen2). As summarized in Table~\ref{tab:model}, these models differ in the number and size of experts, as well as their architectural configurations. Mixtral represents MoE models with a smaller number of larger experts, while Qwen2 and DeepSeek are representative of models with a larger number of smaller experts. Notably, Qwen2 and DeepSeek also incorporate shared experts, which are activated for all input tokens.

\begin{table}
    \centering
\caption{Configuration of three evaluated MoE models.}
\label{tab:model}
    \begin{tabular}{c|c|c|c} \hline \hline
         &  Mixtral&   Qwen2
&DeepSeek\\ \hline 
 \#Layers& 32& 28&26\\ \hline 
         \#Shared Experts&  0&   1
&2\\ \hline 
         \#Routed Experts&  8&   64
&64\\ \hline 
         \#Activated Experts&  2&   8
&6\\ \hline 
         Shared Expert Size&              /&   (3584, 20480)&(2048, 1408)\\ \hline 
         Routed Expert Size&  (4096, 14336)&   (3584, 18944)&(2048, 1408)\\\hline \hline
    \end{tabular}

\end{table}

\subsubsection{Baselines}
We evaluate HybriMoE against three representative open-source MoE inference frameworks: llama.cpp\cite{llamacpp}, AdapMoE\cite{zhong2024adapmoe}, and kTransformers\cite{ktransformers}, each representing a distinct scheduling approach. llama.cpp is a CPU-GPU hybrid scheduling baseline that statically maps model layers to CPU or GPU. AdapMoE is the SOTA for GPU-centric MoE scheduling, minimizing on-demand loading overhead through adaptive prefetching and caching. kTransformers is the SOTA for CPU-GPU hybrid MoE scheduling, mapping high-activation-frequency experts (e.g., shared experts) to the GPU to maximize efficiency.

\subsubsection{Metrics}
Auto-regressive decoding consists of two stages: the prefill stage and the decoding stage. We evaluate the performance of HybriMoE separately for these two stages. For the prefill stage, we use Time To First Token (TTFT), which measures the latency from receiving the input prompt to generating the first token. For the decoding stage, we use Time Between Tokens (TBT), which captures the time taken to generate each subsequent token. These metrics provide a clear assessment of both initial latency and sustained efficiency during inference.

\subsubsection{Datasets}
For the prefill stage, we evaluate performance under varying input lengths by sampling traces of different lengths from multiple datasets, including MT Bench\cite{zheng2024mtbench}, Vicuna Bench\cite{zheng2023judgingllmasajudgemtbenchchatbot} and ChatGPT Prompts\cite{chatgptprompts}. In contrast, for the decoding stage, as performance is not sensitive to input length, we use only the ChatGPT Prompts dataset to evaluate the TBT metric.

\subsection{End-to-End Performance}
\subsubsection{Prefill Stage}
We evaluate the prefill stage performance of HybriMoE by comparing it against three baselines: llama.cpp, AdapMoE, and kTransformers. Figure~\ref{fig:prefill} presents the TTFT results across various input legnths(around 32, 128, 512 and 1024 tokens) and different GPU expert cache ratios(25\%, 50\% and 75\%).

HybriMoE demonstrates consistent improvements over the baselines across all input lengths and cache configurations. llama.cpp exhibits significantly higher prefill latency due to its naive static mapping strategy, which allocates entire layers of experts to the CPU. This approach fails to balance workloads effectively, particularly in the prefill stage where computational demand is high, leading to substantial delays. Compared to kTransformers, HybriMoE achieves an average speedup of \textbf{1.33$\times$} across different input lengths and cache configurations. This improvement is driven by HybriMoE’s hybrid scheduling and impact-driven prefetching mechanisms, which dynamically balance workloads and reduce cache misses, enabling more efficient resource utilization.

\subsubsection{Decode Stage}
Figure~\ref{fig:decode} illustrates the decode performance results for three MoE models.
HybriMoE consistently achieves the highest throughput across all cache ratios and models, demonstrating its ability to dynamically balance workloads and fully utilize hardware resources during the decode stage. Compared to kTransformers, HybriMoE achieves an average throughput improvement of \textbf{1.70$\times$}. Also, it is worth noting that llama.cpp demonstrates relatively strong performance in this stage, especially compared to its prefill stage results. This is primarily due to the smaller computational load per expert in the decode stage, which allows CPU-based computation to proceed faster. Additionally, the impact of uneven expert mapping is less pronounced compared to the prefill stage, and the overall resource overhead remains low, favoring llama.cpp's static scheduling strategy in this specific context.

\subsection{Ablation Study}

\begin{table}
\centering
\caption{MoE inference speedup breakdown of proposed
techniques.}
\label{tab:ablation}
\begin{tblr}{
  cell{2}{1} = {r=4}{},
  cell{6}{1} = {r=5}{},
  vline{2} = {1-10}{},
  vline{3} = {1-10}{},
  hline{1,11} = {-}{},
  hline{2,6} = {-}{},
  colspec={Q[c] Q[c] Q[c] Q[c]},
  abovesep=0.1pt,
  belowsep=0.1pt,
}\hline
        & Technique            & Latency(s) & Speedup \\
Prefill & Baseline             & 1.47        &         \\
        & Baseline+Scheduling  & 1.17        & 1.26$\times$        \\
        & Baseline+Prefetching & 1.39        & 1.06$\times$        \\ \hline
        & All                  & 1.13        & 1.31$\times$        \\
Decode  & Baseline             & 0.21        &         \\
        & Baseline+Scheduling  & 0.14        & 1.46$\times$        \\
        & Baseline+Prefetching & 0.18        & 1.15$\times$        \\
        & Baseline+Caching     & 0.15        & 1.38$\times$        \\ \hline 
        & All                  & 0.11        & 1.86$\times$        \\ \hline
\end{tblr}
\end{table}
We further explore how the components of our method contribute to the result. Performance was measured for Qwen2 under 25\% expert cache ratio for the two stages. The baseline is ktransformers framework. 
The results are illustrated in table~\ref{tab:ablation}.

\subsection{Discussions}


\subsubsection{Score-aware Cache Management Analysis}
\begin{figure}[!tb]
    \centering
    \includegraphics[width=\linewidth]{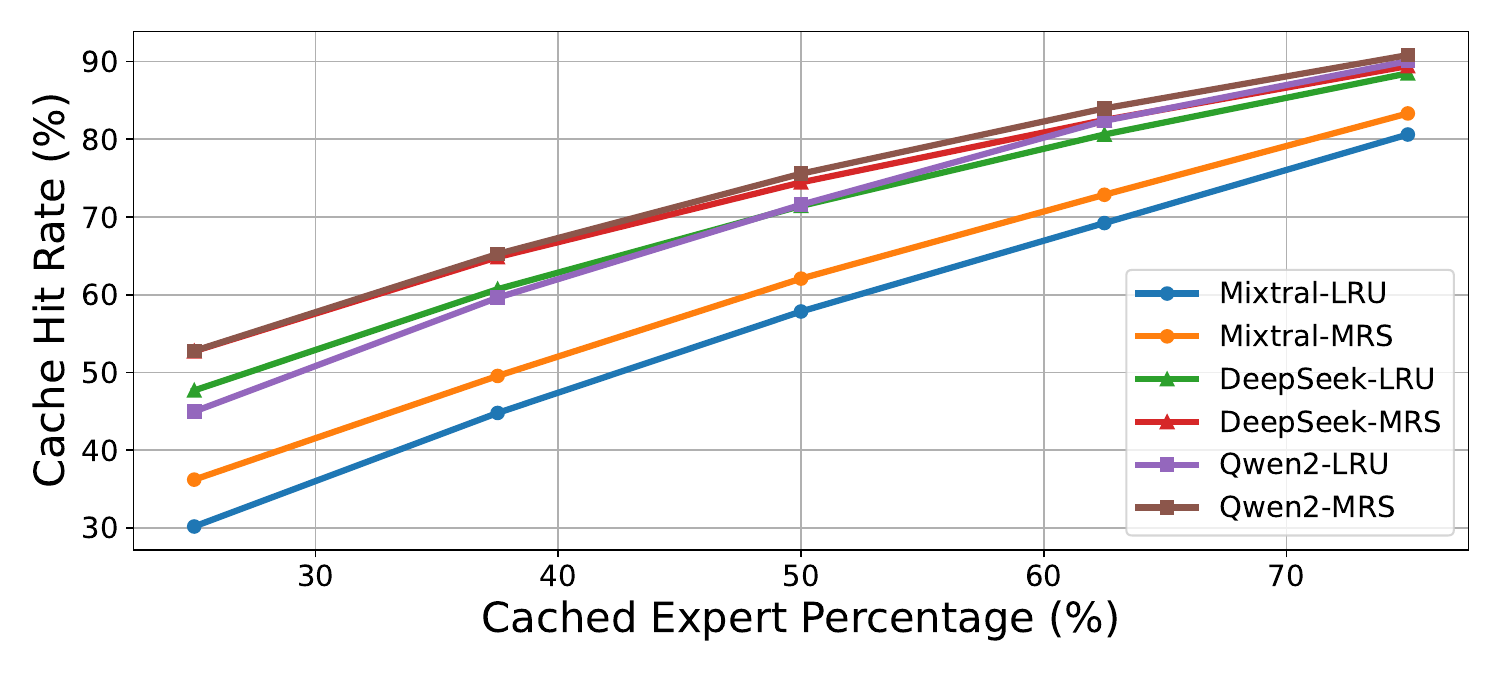}
    \caption{Cache Hit Rate Comparison Between MRS and LRU Across Different Cached Expert Percentages.}
    \label{fig:cache}
\end{figure}
Figure~\ref{fig:cache} compares the cache hit rates of HybriMoE's Minus Recent Score (MRS) strategy and the traditional Least Recently Used (LRU) strategy across three models under varying cached expert percentages. At 25\% cache capacity, MRS outperforms LRU by 6\% to 8\%, with Mixtral improving from 30.2\% to 36.2\%, DeepSeek from 47.7\% to 52.7\%, and Qwen2 from 45.0\% to 52.8\%. As cache capacity increases to 75\%, the gap narrows (e.g., Mixtral: 83.3\% vs. 80.6\%), as higher capacities reduce expert competition, diminishing the relative impact of the caching strategy. These results highlight MRS's effectiveness, particularly under limited cache settings.

\section{CONCLUSION}
This paper presents HybriMoE, a hybrid CPU-GPU scheduling and cache management system designed to address the challenges of MoE inference, including dynamic expert activations and workload imbalances. By incorporating dynamic intra-layer scheduling, impact-driven prefetching, and score-aware caching, HybriMoE achieves efficient resource utilization and reduced latency.
Experiments on various MoE models demonstrate that HybriMoE achieves an average speedup of 1.33x in prefill latency and 1.70x in decode latency compared to state-of-the-art methods. These results highlight HybriMoE’s effectiveness in optimizing hybrid MoE inference and its potential for scalable deployment on resource-constrained devices.
\newpage
\bibliographystyle{IEEEtran}
\bibliography{reference/schedule.bib}

\end{document}